\pdfoutput=1

\documentclass[11pt]{article}

\usepackage[final]{acl}

\usepackage{times}
\usepackage{latexsym}

\usepackage[T1]{fontenc}

\usepackage[utf8]{inputenc}

\usepackage{microtype}

\usepackage{inconsolata}

\usepackage{graphicx}

\usepackage{inconsolata}
\usepackage{enumerate}
\usepackage{enumitem}
\usepackage{tikz}
\usepackage{amsmath}
\usepackage{subcaption}
\usepackage{amssymb}
\usepackage{booktabs}
\usepackage{arydshln}
\usepackage{xcolor}
\definecolor{customorange}{RGB}{237,125,49}
\definecolor{customblue}{RGB}{68,114,196}
\definecolor{customgreen}{RGB}{0,176,80}
\usepackage[normalem]{ulem}
\useunder{\uline}{\ul}{}
\definecolor{royalblue}{rgb}{0.25, 0.41, 0.88}

\usepackage{ulem}
\usepackage{todonotes}

\title{Improving Fairness of Large Language Models in Multi-document Summarization}

\author{
  \textbf{Haoyuan Li\textsuperscript{1}},
  \textbf{Rui Zhang\textsuperscript{2}},
  \textbf{Snigdha Chaturvedi\textsuperscript{1}}
\\
  \textsuperscript{1}University of North Carolina at Chapel Hill,
  \textsuperscript{2}Pennsylvania State University 
\\
  {\{haoyuanl, snigdha\}@cs.unc.edu, \{ rmz5227\}@psu.edu}
}

\begin{document}
\maketitle
\begin{abstract}
Fairness in multi-document summarization (MDS) is crucial for providing comprehensive views across documents with diverse social attribute values, which can significantly impact decision-making. For example, a summarization system that tends to overrepresent negative reviews of products can mislead customers into disregarding good products. Previous works measure fairness in MDS at two levels: summary-level and corpus-level. While summary-level fairness focuses on individual summaries, corpus-level fairness focuses on a corpus of summaries. Recent methods primarily focus on summary-level fairness. We propose FairPO, a preference tuning method that focuses on both summary-level and corpus-level fairness in MDS. To improve summary-level fairness, we propose to generate preference pairs by perturbing document sets. To improve corpus-level fairness, we propose fairness-aware preference tuning by dynamically adjusting the weights of preference pairs. Our experiments show that FairPO outperforms strong baselines while maintaining the critical qualities of summaries. The code is available at \href{https://github.com/leehaoyuan/coverage_fairness}{https://github.com/leehaoyuan/coverage\_fairness}.
\end{abstract}

\section{Introduction}
\label{intro}


Multi-document summarization (MDS) aims to summarize the salient information from multiple documents about an entity, such as  reviews of a product. 
Each of these documents is generally associated with a \textit{social attributes} such as sentiments in reviews. These documents with different social attribute values e.g. positive sentiment or negative sentiment tend to have diverse information or conflicting opinions. It is crucial that the summary fairly represents conflicting information since it can significantly impact decision-making. 

Previous works \cite{shandilya2018fairness,  olabisi-etal-2022-analyzing, huang2024bias} measure fairness in MDS at two levels: summary-level or corpus-level. Summary-level fairness measures how fairly a summary represents documents with different social attribute values. Corpus-level fairness measures how fairly a corpus of summaries as a whole represents different social attribute values. 

Recent studies \cite{zhang2023fair, li2024cover} find that modern summarization methods like LLMs struggle with both summary-level and corpus-level fairness. To improve the summary-level fairness, \citet{zhang2023fair} prompt LLMs to generate summaries based on the distribution of social attributes among documents. However, it relies on users' prior knowledge of fairness issues and social attributes, limiting its effectiveness in practice. \citet{huang2024bias} improve the summary-level fairness of T5 \cite{raffel2020exploring} by policy gradient, but their method may not generalize to modern models like LLMs. Furthermore, both methods focus exclusively on summary-level fairness, overlooking the corpus-level fairness. 

 We propose FairPO (\textbf{F}air \textbf{P}reference \textbf{O}ptim-ization), a preference tuning \cite{ziegler2019fine} method that focuses on both summary-level and corpus-level fairness of LLMs in MDS. While previous works \cite{stiennon2020learning, roit-etal-2023-factually} uses preference tuning to improve other qualities of summaries, FairPO is the first to use preference tuning for the fairness in MDS.  FairPO is based on Direct Preference Optimization  (DPO) \cite{rafailov2024direct}. 
To optimize summary-level fairness, FairPO generates preference pairs given perturbed input document sets by removing a small subset of documents with certain social attribute values. To further improve corpus-level fairness, FairPO performs \textit{fairness-aware preference tuning} by dynamically adjusting the weights of preference pairs. 

We conduct an empirical evaluation of FairPO using three LLMs: Llama3.1 \cite{llama3modelcard}, Mistral \cite{jiang2023mistral}, and Gemma2 \cite{team2024gemma}, on the Amazon \cite{ni-etal-2019-justifying}, MITweet \cite{liu2023ideology}, and SemEval datasets \cite{mohammad-etal-2016-semeval}. Our experiments show that FairPO outperforms strong baselines while maintaining other critical qualities of summaries, such as relevance and factuality. 

Our contributions are as follows: 
\begin{itemize}[topsep=1pt, leftmargin=*, noitemsep]
    \itemsep0mm
    \item We propose FairPO to improve the fairness of LLMs in MDS;  
    \item We propose to improve summary-level 
 and corpus-level fairness by perturbation-based preference pair generation and fairness-aware preference tuning;  
    \item We perform comprehensive experiments to show the effectiveness of FairPO. 
\end{itemize}



\section{Background}

In this section, we provide background knowledge on fairness in MDS. Let $G$ denote all document sets in a corpus for MDS. Each document set $D \in G$ contains multiple documents $\{d_1,...,d_n\}$, where each document $d_i$ is labeled with a social attribute $a_i \in \{1,...,K\}$. For each document set $D$, a MDS system is supposed to generate a summary $S$.

To evaluate fairness in MDS, we use Equal Coverage $EC(D,S)$, a summary-level measure, and Coverage Parity $CP(G)$, a corpus-level measure, proposed by \citet{li2024cover}. Below, we summarize these concepts as introduced in the original paper.

\paragraph{Equal Coverage} examines whether each social attribute value has equal probabilities of being covered by the summary $S$ for a document set $D$. Specifically, it first defines \textbf{coverage probability difference} $c(d_i,S)$ as the difference between the coverage probability for the document $d_i$, $p(d_i,s)$. It also defines the average coverage probability across all documents, 
$p(d,s)$. To estimate the coverage probability for the document $d_i$, $p(d_i,s)$, FairPO estimates the probability $p(d_i,s_j)$ that a document $d_i$ is covered by a summary sentence $s_j$. Specifically, the probability $p(d_i,s_j)$ is estimated as the maximum entailment probability $p(d_{i,l}, s_j)$ between any document chunk $d_{i,l}$ of the document $d_i$ and the summary sentence $s_j$ using an entailment model: 
\vspace{-0.15cm}
\begin{equation}
p(d_i,s_j)=\max\{p(d_{i,l},s_j)|d_{i,l}\in d_i\},
\label{equ:entail}
\end{equation}
The coverage probability for the document $d_i$, $p(d_i,s)$, is then estimated as the average of the probability $p(d_i,s_j)$:
\begin{equation}
\vspace{-0.15cm}
p(d_i,s)=\frac{1}{|S|} \sum_{s_j\in S} p(d_i,s_j),
\vspace{-0.05cm}
\end{equation}
The average coverage probability, $p(d,s)$, is then calculated by averaging coverage probability, $p(d_i,s)$, across all documents in the document set $D$. Using these values, Equal Coverage calculates the coverage probability difference $c(d_i,S)=p(d_i,s)-p(d,s)$. Equal Coverage value $EC(D,S)$ is then calculated as the average of the absolute average coverage probability difference $c(d_i,S)$ for documents with each social attribute value:
\begin{equation}
\vspace{-0.15cm}
EC(D,S)=\frac{1}{K}\sum_{k=1}^{K}|\mathbb{E}(\{c(d_i,S)|a_i=k\})|
\vspace{-0.05cm}
\end{equation}
A lower $EC(D,S)$ indicates a fairer summary $S$. To evaluate the fairness of a system, we use the average Equal Coverage value across the corpus $G$.

\paragraph{Coverage Parity} examines whether certain social attribute values are systematically overrepresented or underrepresented across the corpus $G$. Coverage Parity collects these coverage probability differences $c(d_i,S)$ from all input documents of the dataset $G$ whose social attribute value is $k$ into a set $C_k$. The coverage Parity value $CP(G)$ is then calculated as the average of the absolute average coverage probability difference $c(d_i,S)$ for documents with each social attribute value:
\begin{equation}
\vspace{-0.2cm}
CP(G)= \frac{1}{K}\sum_{k=1}^{K}|\mathbb{E}(C_k)|,
\vspace{-0.05cm}
\label{equ:cp}
\end{equation}
A lower $CP(G)$ indicates a fairer system. For more details, please refer to \citet{li2024cover}.

\section{FairPO} 
In this section, we describe our proposed preference tuning method, FairPO.
\subsection{Perturbation-based Preference Pair Generation}
\label{sec:pertub}
In this section, we describe how to generate preference pairs based on perturbation. A preference pair for FairPO contains a chosen summary $S_c$ and a rejected summary $S_r$ for the document set $D$. Ideally, the chosen and rejected summaries should differ significantly in representing documents with different social attribute values. To this end, FairPO generates summaries for perturbed input document sets, where small subsets ($\alpha$\%) of documents with specific social attribute values are removed.  

Specifically, FairPO first generates a summary $S$ for the input document set $D$ and identifies its most overrepresented, $k^+$, and underrepresented, $k^-$, social attribute value. For the completeness of information, FairPO only considers social attribute values that appear in more than $\alpha\%$ of the documents (details in App. \ref{sec:implementation}). These are determined based on the highest or lowest average coverage probability differences, $\mathbb{E}(\{c(d_i,S)|a_i=k\})$. Then, FairPO generates summary $S^+$ and $S^-$ for the perturbed input document set where $\alpha\%$ of randomly sampled documents with social attribute value $a_i$ of $k^+$ and $k^-$ are removed. Among summaries $S$, $S^+$, $S^-$, FairPO selects the summary with the lowest Equal Coverage value, indicating the best summary-level fairness, as the chosen summary $S_c$. The summary with the highest Equal Coverage value is selected as the rejected summary $S_r$.

\subsection{Fairness-aware Preference Tuning}
\label{sec:weight}
In this section, we describe fairness-aware preference tuning that optimizes summary-level and corpus-level fairness. To achieve this, FairPO dynamically assigns separate weights for the chosen summary $S_c$ and the rejected summary $S_r$ based on estimated corpus-level fairness during training.

FairPO modifies the DPO objective (more explanations in App. \ref{sec:relation}) and introduces separate weights, $w_c$ and $w_r$, for the chosen summary $S_c$ and rejected summary $S_r$ respectively:
\vspace{-0.1cm}
\begin{equation}
\sigma(-m)\beta(w_rlog\frac{\pi_\theta(S_r|D)}{\pi_{ref}(S_r|D)}-w_c log\frac{\pi_\theta(S_c|D)}{\pi_{ref}(S_c|D)})
\label{eq:objective}
\end{equation}
\vspace{-0.1cm}where $\sigma$ is the sigmoid function, $\pi_\theta$ is the policy model, $\pi_{ref}$ is the reference model, and $m$ is the reward margin as in DPO:
\vspace{-0.2cm}
\begin{equation}
m=\beta log\frac{\pi_\theta(S_c|D)}{\pi_{ref}(S_c|D)}-\beta log\frac{\pi_\theta(S_r|D)}{\pi_{ref}(S_r|D)}
\end{equation}
\vspace{-0.1cm}The term $\sigma(-m)$ in Eq. \ref{eq:objective} serves as a scaling factor and FairPO does not consider its gradient.

FairPO assigns weights $w_c$ and $w_r$ to summaries based on their impact on corpus-level fairness. It assigns high weights $w_c$ to chosen summaries that improve corpus-level fairness by balancing the overrepresentation and underrepresentation of social attribute values. Conversely, it assigns high weights $w_r$ to rejected summaries that hurt corpus-level fairness. To estimate corpus-level fairness, FairPO computes the sum of coverage probability differences for documents with social attribute values of $k$, $C_k(D,S_*)=\sum_{d\in \{d_i|a_i=k\}} c(d,S_*)$ 
for each chosen or rejected summary, $S_*$. A summary $S_*$ is considered overrepresenting or underrepresenting the social attribute value $k$ if the sum of coverage probability differences, $C_k(D,S_*)$, is greater or less than zero respectively. In each training step, FairPO estimates the overrepresentation $O(k)$ of social attribute value $k$:
\vspace{-0.1cm}
\begin{equation}
O(k)=\frac{\sum_{(D,S)\in T_k^+}|C_k(D,S)|\cdot\pi_\theta(S|D)/|S|}{\sum_{(D,S)\in T_k^+}\pi_\theta(S|D)/|S|}
\label{eq:over}
\end{equation}
\vspace{-0.1cm}\noindent where $T_k^+$ is the set of document sets $D$ and corresponding chosen or rejected summaries that overrepresent social attribute value $k$ ($C_k(D,S_*)>0$) in recent training steps. 
Similarly, FairPO estimates the underrepresentation $U(k)$ using the set $T^-_k$ of document sets and summaries that underrepresent social attribute value $k$ ($C_k(D,S_*)<0$) as Eq. \ref{eq:over}. 

Using the overrepresentation $O(k)$ and underrepresentation $U(k)$, FairPO assigns weight $w_c$ and $w_r$. Chosen summaries that help balance overrepresentation $O(k)$ and underrepresentation $U(k)$ receive higher weights and vice versa for rejected summaries. For example, the weight $w_c$ should be higher if a systematically underrepresented social attribute value $k$ ($U(k)>O(k)$) is overrepresented by the chosen summary $S_c$ ($C_k(D,S_c)>0$). For social attribute value $k$, FairPO computes an intermediate weight $w_{c,k}$ for the chosen summary $S_c$:
\vspace{-0.1cm}
\begin{equation}
w_{c,k}=\frac{2}{1+(O(k)/(U(k))^{C_k(D,S_c)/\tau}}
\label{eq:cond1}
\end{equation}
\vspace{-0.1cm}where $\tau$ is the temperature. The weight $w_c$ for chosen summaries is the average intermediate weight $w_{c,k}$ across all social attribute values. The weight $w_r$ for the rejected summary $S_r$ is computed similarly with the intermediate weight $w_{r,k}$:
\vspace{-0.1cm}
\begin{equation}
w_{r,k}=\frac{2}{1+(U(k)/(O(k))^{C_k(D,S_r)/\tau}}
\label{eq:cond2}
\end{equation}
\vspace{-0.1cm}The design ensures that summaries improving corpus-level fairness are prioritized.



\begin{table}[t]
\centering
\small
\setlength{\tabcolsep}{1mm}
\resizebox{0.48\textwidth}{!}{
\begin{tabular}{lccccc}
\toprule
                        & Domain & Soci. Attr. & Soci. Attr. Val.                                                      & Doc. Set Size & Doc. Len \\
\midrule
    
Amazon  & Review & Sentiment   & \begin{tabular}[c]{@{}c@{}}negative, neutral, \\ positive\end{tabular} & 8             & 40       \\
MiTweet & Tweet  & Ideology    & \begin{tabular}[c]{@{}c@{}}left, center,\\ right\end{tabular}          & 20            & 34       \\
SemEval & Tweet  & Stance      & support, against                                                       & 30            & 17      \\
\bottomrule
\end{tabular}}
\caption{Dataset statistics. Doc. Set Size means size of document sets. Doc. Len. means average length of documents.}
\label{tab:dataset}
\end{table}


\section{Experiments} 
In this section, we describe experiments of finetuning models with FairPO.
\subsection{Datasets}
\label{sec:dataset_main}
We experiment on three datasets: Amazon \cite{ni-etal-2019-justifying}, MITweet \cite{liu2023ideology}, SemEval \cite{mohammad-etal-2016-semeval} datasets. Each dataset includes $1000$ samples for training, $300$ samples for validation, and $300$ samples for testing. The division of training, validation, and testing sets is based on stratified sampling of social attribute values and topics. Tab. \ref{tab:dataset} shows the statistics of these datasets. The summary length is 50 words. Details of preprocessing are in App. \ref{sec:dataset}.

\subsection{Implementation Details}
We perform experiments with three LLMs: Llama3.1-8b-Instruct \cite{llama3modelcard} , Mistral-7B-Instruct-v0.3 \cite{jiang2023mistral}, Gemma-2-9b-it \cite{team2024gemma}. Each LLM is trained for 2 epochs using LoRA \cite{hulora} with a learning rate of $5e-5$ and batch size of $16$. To generate preference pairs, FairPO removes $\alpha=10\%$ of documents. The temperature $\tau$ is $1$ on the MITweet dataset, $2$ for Mistral and $1$ for other LLMs on the Amazon dataset, $3$ for Mistral and $2$ for other LLMs on the SemEval dataset. All hyperparameters are tuned on the validation set. More details are in App. \ref{sec:implementation}.


\begin{table}[]
\setlength{\tabcolsep}{1mm}
\resizebox{0.48\textwidth}{!}{
\begin{tabular}{lcccccccc}
\hline

                 & \multicolumn{2}{c}{Amazon}                & \multicolumn{2}{c}{MITweet}         & \multicolumn{2}{c}{SemEval}         &  \multicolumn{2}{c}{Overall}               \\
                 & $EC\downarrow$                  & $CP\downarrow$                  & $EC$                  & $CP\downarrow$            & $EC\downarrow$                  & $CP\downarrow$            & $\overline{EC}\downarrow$           & $\overline{CP}\downarrow$            \\ \hline
Llama3.1   & 7.95          & 1.89          & 4.50          & 0.59          & 2.98          & 1.41          & 5.14          & 1.30          \\
~+DPO       & 7.23          & 1.27          & {\ul 4.25}    & 0.47          & 2.66          & 1.09          & 4.72          & 0.94          \\
~+OPTune    & \textbf{6.70} & {\ul 0.62}    & 4.33          & 0.51          & {\ul 2.60}    & 0.95          & {\ul 4.54}    & {\ul 0.69}    \\
~+Prompt    & 7.42          & 1.64          & 4.36          & {\ul 0.45}    & 2.62          & \textbf{0.29} & 4.80          & 0.79          \\
~+Policy G. & 7.73          & 1.88          & 4.51          & 0.55          & 2.97          & 1.38          & 5.07          & 1.27          \\
~+FairPO    & {\ul 6.87}    & \textbf{0.42} & \textbf{4.24} & \textbf{0.42} & \textbf{2.49} & {\ul 0.66}    & \textbf{4.53} & \textbf{0.50} \\ \hdashline
Mistral    & 8.36          & 2.83          & 4.16          & 0.61          & 2.83          & 1.27          & 5.12          & 1.57          \\
~+DPO       & 7.20          & 1.82          & \textbf{3.55} & \textbf{0.34} & 2.41          & 0.93          & 4.39          & 1.03          \\
~+OPTune    & {\ul 6.85}          & {\ul 0.88}          & {\ul 3.58}    & 0.51          & \textbf{2.07} & 0.57          & {\ul 4.17}    & {\ul 0.65}    \\
~+Prompt    & 7.74          & 1.92          & 3.97          & {\ul 0.37}    & 2.35          & \textbf{0.36} & 4.68          & 0.88          \\
~+FairPO    & \textbf{6.32} & \textbf{0.46} & 3.70          & 0.40          & {\ul 2.10}    & {\ul 0.43}    & \textbf{4.04} & \textbf{0.43} \\ \hdashline
Gemma2     & 8.32          & 2.48          & 4.20          & 0.60          & 2.81          & 0.96          & 5.11          & 1.35          \\
~+DPO       & 6.90          & 0.91          & 4.04          & {\ul 0.40}    & {\ul 2.44}    & 0.56          & 4.46          & {\ul 0.62}    \\
~+OPTune    & {\ul 6.84}    & {\ul 0.88}    & {\ul 3.89}    & 0.57          & \textbf{2.32} & 0.49          & {\ul 4.35}    & 0.65          \\
~+Prompt    & 7.28          & 1.16          & 4.33          & \textbf{0.32} & 2.73          & {\ul 0.48}    & 4.78          & 0.65          \\
~+FairPO    & \textbf{6.18} & \textbf{0.44} & \textbf{3.76} & 0.48          & 2.50          & \textbf{0.45} & \textbf{4.15} & \textbf{0.46} \\
 \hline
\end{tabular}}
\caption{Summary-level fairness ($EC$) and corpus-level fairness ($CP$) of summaries generated by different methods. The best performing method is in \textbf{bold}. The second-best performing method is \underline{underlined}. FairPO has the best overall performance.}
\label{tab:result}
\end{table}

\subsection{Automatic Evaluation of FairPO}
We compare FairPO with the following baselines: (i) DPO \cite{rafailov2024direct}, where the chosen and rejected summaries are selected among three randomly sampled summaries based on EC values like FairPO for a fair comparison; (ii) OPTune \cite{chen2024optune}, which selects the chosen and rejected summaries as DPO and weights preference pairs based on EC value differences; (iii) Policy gradients \cite{lei2024polarity} and (iv) a prompting method \cite{zhang2023fair}. Implementation details of these baselines are in App \ref{sec:baseline}. For evaluation, we consider summary-level and corpus-level fairness using Equal Coverage (EC) and Coverage Parity (CP) \cite{li2024cover}. A lower value is better for these measures. We report the average results on three splittings of training, validation and testing in Tab. \ref{tab:result}.We additionally report the results for each splitting in App. \ref{sec:split}. We observe that FairPO outperforms other methods for most LLMs on all datasets and yields the best overall performance for all LLMs. The results show that FairPO improves both summary-level and corpus-level fairness. 



\subsection{Ablation Study}
\label{sec:ablation}
To validate the effect of perturbation-based preference pair generation and fairness-aware preference tuning, we compare FairPO with its ablated versions. We consider the following ablated versions: (i) (w/o pert.), where the chosen and rejected summaries are selected among three randomly sampled summaries based on Equal coverage values; (ii) (w/o fair.) that performs preference tuning using the DPO objective instead of the fairness-aware preference tuning; (iii) (w/o rew.) that directly assigns weights $w_c$ and $w_r$ in the DPO objective (Eq. \ref{eq:reward}), which undermines the effectiveness of reward margin (more explanations in App. \ref{sec:relation}). Tab.\ref{tab:ablat} reports the results for each dataset,  and \textit{Overall} scores, which is the average across all datasets. A lower value indicates better fairness. 
\begin{table}[]
\setlength{\tabcolsep}{1mm}
\resizebox{0.48\textwidth}{!}{
\begin{tabular}{lcccccccc}
\hline

                 & \multicolumn{2}{c}{Amazon}                & \multicolumn{2}{c}{MITweet}         & \multicolumn{2}{c}{SemEval}         &  \multicolumn{2}{c}{Overall}               \\
                 & $EC\downarrow$                  & $CP\downarrow$                  & $EC$                  & $CP\downarrow$            & $EC\downarrow$                  & $CP\downarrow$            & $\overline{EC}\downarrow$           & $\overline{CP}\downarrow$            \\ \hline
\multicolumn{9}{c}{Llama3.1}                                                                                                                             \\
FariPO                   & 6.57          & \textbf{0.37} & 4.20          & \textbf{0.26} & 2.39          & \textbf{0.56} & \textbf{4.39} & \textbf{0.39} \\
~w/o pert. & 7.01          & 0.48          & \textbf{4.07} & 0.34          & 2.54          & 0.81          & 4.54          & 0.54          \\
~w/o fair. & 6.70          & 0.95          & 4.26          & 0.31          & \textbf{2.29} & 0.65          & 4.42          & 0.64          \\
~w/o rew   & \textbf{6.48} & 0.79          & 4.19          & 0.27          & 2.60          & 0.86          & 4.42          & 0.64          \\ \hdashline
\multicolumn{9}{c}{Mistral}                                                                                                                              \\
FariPO                   & \textbf{6.98} & \textbf{0.89} & \textbf{3.56} & 0.21          & \textbf{1.97} & \textbf{0.36} & \textbf{4.17} & \textbf{0.49} \\
~w/o pert. & 7.29          & 1.64          & 3.81          & 0.21          & 2.30          & 0.26          & 4.47          & 0.71          \\
~w/o fair. & 7.31          & 1.36          & 3.57          & 0.25          & 2.21          & 0.66          & 4.37          & 0.76          \\
~w/o rew   & 7.05          & 1.26          & 3.65          & \textbf{0.14} & 2.06          & 0.55          & 4.25          & 0.65          \\ \hdashline
\multicolumn{9}{c}{Gemma2}                                                                                                                               \\
FariPO                   & \textbf{6.09} & \textbf{0.33} & \textbf{3.84} & 0.47          & 2.53          & 0.59          & \textbf{4.15} & 0.46          \\
~w/o pert. & 6.18          & 0.19          & 4.17          & \textbf{0.21} & 2.43          & 0.53          & 4.26          & \textbf{0.31} \\
~w/o fair. & 6.77          & 1.11          & 3.84          & 0.51          & \textbf{2.39} & 0.59          & 4.34          & 0.74          \\
~w/o rew   & 6.89          & 0.90          & 3.94          & 0.40          & 2.49          & \textbf{0.44} & 4.44          & 0.58          \\
 \hline
\end{tabular}}
\caption{Summary-level fairness ($EC$) and corpus-level fairness ($CP$) of summaries generated by ablated versions of FairPO. The best performing method is in \textbf{bold}.  FairPO has the best overall performance.}
\label{tab:ablat}
\end{table}

From the table, we observe that FairPO yields the best overall performance compared to its ablated versions. The results show the effectiveness of perturbation-based preference pair generation and fairness-aware preference tuning. It also provides empirical evidences for the design choice of objective of FairPO.

\subsection{Human Evaluation of FairPO}
\label{sec:human_eval}
We perform a human evaluation to compare the fairness of summaries generated by LLMs tuned with DPO and FairPO. For each LLM, we randomly select $10$ pairs of summaries generated by the LLM tuned with DPO or FairPO, yielding a total of $30$ pairs. Each pair is annotated by three annotators recruited from Amazon Mechanical Turk. Annotators are asked to read all corresponding documents and select the fairer summary. We choose the Amazon dataset since each document set only contains eight reviews (Tab. \ref{tab:dataset}) and judging the sentiment of an opinion is relatively easy for common users. The Randolph’s Kappa \cite{randolph2005free} between annotations of three annotators is $0.40$, which shows a moderate correlation. The correlation is expected considering the subjectivity of the task. More details are in App. \ref{app:human}.

Out of $30$ pairs, summaries generated by FairPO-tuned LLMs are fairer in $18$ pairs and summaries generated by DPO-tuned LLMs are fairer in $9$ pairs.The difference is statistically significant ($p<0.05$) using bootstrap \cite{koehn-2004-statistical}. The results show that FairPO performs better than DPO in improving fairness. We additionally show example summaries generated by FairPO in App. \ref{sec:qualitative}. 
\begin{table}[]
\setlength{\tabcolsep}{0.75mm}
\resizebox{0.48\textwidth}{!}{
\begin{tabular}{lccccccccc}
\hline
       & \multicolumn{3}{c}{Llama3.1}   & \multicolumn{3}{c}{Mistral}       & \multicolumn{3}{c}{Gemma2}                  \\
       & flu.$\uparrow$       & rel.$\uparrow$      & fac.$\uparrow$ & flu.$\uparrow$  & rel.$\uparrow$        & fac.$\uparrow$        & flu.$\uparrow$         & rel.$\uparrow$         & fac.$\uparrow$         \\ \hline
DPO    & {\ul 7.56} & {\ul 8.33} & 2.78 & 5.11  & {\ul 11.56} & {\ul 11.56} & {\ul 5.11}   & 1.11         & {\ul 8.67}   \\
OPTune &   1.00         &    0.44        &  {\ul -6.89}    & -0.78 & {\ul 6.78}  & {\ul 8.89}  & {\ul 7.00}   & {\ul 11.67}  & {\ul 11.67}  \\
Prompt &   {\ul -15.33}         &   {\ul -19.22}         & {\ul -24.44}     & -0.44 & {\ul -6.00} & {\ul -5.56} & {\ul -42.67} & {\ul -50.78} & {\ul -51.44} \\
FairPO & {\ul 5.78} & 3.11       & 2.89 & 2.11  & {\ul 5.33}  & {\ul 9.11}  & {\ul 11.44}  & {\ul 16.11}  & {\ul 9.44}   \\ \hline
\end{tabular}}
\caption{Pairwise comparison of quality between summaries generate by LLMs before and after tuning. Statistical significant differences ($p<0.05$) according to paired bootstrap resampling \cite{koehn-2004-statistical} are underlined. FairPO does not affect summary quality. }
\label{tab:quality}
\end{table}

\subsection{Evaluation of Summary Quality}

To evaluate FairPO's impact on summary quality, we compare summaries generated by LLMs before and after tuning to improve fairness. 
Specifically, for a pair of summaries, we instruct Prometheus 2 (7B) \cite{kim2024prometheus}  to select the better summary in three dimensions: fluency, relevance, and factuality. To mitigate position bias \cite{huang2023embrace}, we perform the pairwise comparison twice with different orders of summaries and only consider consistent results. Tab. \ref{tab:quality} reports the differences between the winning and losing rates of different methods. A positive value indicates summary quality is better compared to original LLMs.

From the table, we observe that the quaility of summaries generated by LLMs tuned with FairPO is comparable with summaries generated by original LLMs. Contrarily, prompting significantly hurt the quality of summaries. The results show that FairPO improves the fairness of summaries while maintaining their quality. 

\section{Conclusion}
We propose FairPO, a preference tuning method that optimizes summary-level fairness and corpus-level fairness in MDS. Specifically, FairPO generates preference pairs using perturbed document sets to improve summary-level fairness and performs fairness-aware preference tuning to improve corpus-level fairness. Our experiments show that FairPO outperforms strong baselines while maintaining critical qualities of summaries.

\section{Limitation}
Our experiments demonstrate FairPO's effectiveness in improving both summary-level and corpus-level fairness of summaries within individual domains. While this work focuses on optimizing fairness within a single domain, extending FairPO to improve fairness simultaneously across multiple domains with diverse social attributes presents a promising future direction. Besides, FairPO currently selects the two summaries with the largest fairness differences among the three generated summaries for preference tuning, following commonly used practices of DPO. Exploring approaches to utilize all three summaries generated by FairPO can be another interesting future direction.

\section{Acknowledgment}
This work was supported by NSF grant DRL-2112635 and 2338418.

\section{Ethical Consideration}
The datasets we use are all publicly available. We do not annotate any data on our own. All the models used in this paper are publicly accessible. The inference and finetuning of models are performed on one Nvidia A6000 or Nvidia A100 GPU.

We perform human evaluation experiments on Amazon Mechanical Turk. The annotators were compensated at a rate of \$20 per hour. During the evaluation, human annotators were not exposed to any sensitive or explicit content.
\bibliography{acl_latex}

\begin{thebibliography}{24}
\providecommand{\natexlab}[1]{#1}

\bibitem[{AI@Meta(2024)}]{llama3modelcard}
AI@Meta. 2024.
\newblock \href {https://github.com/meta-llama/llama3/blob/main/MODEL_CARD.md} {Llama 3 model card}.

\bibitem[{Chen et~al.(2024)Chen, Chen, Liu, Kirchenbauer, Soselia, Zhu, Goldstein, Zhou, and Huang}]{chen2024optune}
Lichang Chen, Jiuhai Chen, Chenxi Liu, John Kirchenbauer, Davit Soselia, Chen Zhu, Tom Goldstein, Tianyi Zhou, and Heng Huang. 2024.
\newblock Optune: Efficient online preference tuning.
\newblock \emph{arXiv preprint arXiv:2406.07657}.

\bibitem[{Hu et~al.(2021)Hu, Wallis, Allen-Zhu, Li, Wang, Wang, Chen et~al.}]{hulora}
Edward~J Hu, Phillip Wallis, Zeyuan Allen-Zhu, Yuanzhi Li, Shean Wang, Lu~Wang, Weizhu Chen, et~al. 2021.
\newblock Lora: Low-rank adaptation of large language models.
\newblock In \emph{International Conference on Learning Representations}.

\bibitem[{Huang et~al.(2023)Huang, Laban, Fabbri, Choubey, Joty, Xiong, and Wu}]{huang2023embrace}
Kung-Hsiang Huang, Philippe Laban, Alexander~R Fabbri, Prafulla~Kumar Choubey, Shafiq Joty, Caiming Xiong, and Chien-Sheng Wu. 2023.
\newblock Embrace divergence for richer insights: A multi-document summarization benchmark and a case study on summarizing diverse information from news articles.
\newblock \emph{arXiv preprint arXiv:2309.09369}.

\bibitem[{Huang et~al.(2024)Huang, Fayek, and Zhang}]{huang2024bias}
Nannan Huang, Haytham Fayek, and Xiuzhen Zhang. 2024.
\newblock Bias in opinion summarisation from pre-training to adaptation: A case study in political bias.
\newblock \emph{arXiv preprint arXiv:2402.00322}.

\bibitem[{Jiang et~al.(2023)Jiang, Sablayrolles, Mensch, Bamford, Chaplot, Casas, Bressand, Lengyel, Lample, Saulnier et~al.}]{jiang2023mistral}
Albert~Q Jiang, Alexandre Sablayrolles, Arthur Mensch, Chris Bamford, Devendra~Singh Chaplot, Diego de~las Casas, Florian Bressand, Gianna Lengyel, Guillaume Lample, Lucile Saulnier, et~al. 2023.
\newblock Mistral 7b.
\newblock \emph{arXiv preprint arXiv:2310.06825}.

\bibitem[{Kim et~al.(2024)Kim, Suk, Longpre, Lin, Shin, Welleck, Neubig, Lee, Lee, and Seo}]{kim2024prometheus}
Seungone Kim, Juyoung Suk, Shayne Longpre, Bill~Yuchen Lin, Jamin Shin, Sean Welleck, Graham Neubig, Moontae Lee, Kyungjae Lee, and Minjoon Seo. 2024.
\newblock \href {https://arxiv.org/abs/2405.01535} {Prometheus 2: An open source language model specialized in evaluating other language models}.
\newblock \emph{Preprint}, arXiv:2405.01535.

\bibitem[{Koehn(2004)}]{koehn-2004-statistical}
Philipp Koehn. 2004.
\newblock \href {https://aclanthology.org/W04-3250} {Statistical significance tests for machine translation evaluation}.
\newblock In \emph{Proceedings of the 2004 Conference on Empirical Methods in Natural Language Processing}, pages 388--395, Barcelona, Spain. Association for Computational Linguistics.

\bibitem[{Lei et~al.(2024)Lei, Song, Cho, Wang, Huang, and Yu}]{lei2024polarity}
Yuanyuan Lei, Kaiqiang Song, Sangwoo Cho, Xiaoyang Wang, Ruihong Huang, and Dong Yu. 2024.
\newblock Polarity calibration for opinion summarization.
\newblock \emph{arXiv preprint arXiv:2404.01706}.

\bibitem[{Li et~al.(2024)Li, Zhang, Zhang, and Chaturvedi}]{li2024cover}
Haoyuan Li, Yusen Zhang, Rui Zhang, and Snigdha Chaturvedi. 2024.
\newblock \href {https://arxiv.org/abs/2412.08795} {Coverage-based fairness in multi-document summarization}.
\newblock \emph{Preprint}, arXiv:2412.08795.

\bibitem[{Liu et~al.(2023)Liu, Luo, Xu, Wei, Wei, Yu, Xiang, and Wang}]{liu2023ideology}
Songtao Liu, Ziling Luo, Minghua Xu, LiXiao Wei, Ziyao Wei, Han Yu, Wei Xiang, and Bang Wang. 2023.
\newblock Ideology takes multiple looks: A high-quality dataset for multifaceted ideology detection.
\newblock In \emph{The 2023 Conference on Empirical Methods in Natural Language Processing}.

\bibitem[{Mohammad et~al.(2016)Mohammad, Kiritchenko, Sobhani, Zhu, and Cherry}]{mohammad-etal-2016-semeval}
Saif Mohammad, Svetlana Kiritchenko, Parinaz Sobhani, Xiaodan Zhu, and Colin Cherry. 2016.
\newblock \href {https://doi.org/10.18653/v1/S16-1003} {{S}em{E}val-2016 task 6: Detecting stance in tweets}.
\newblock In \emph{Proceedings of the 10th International Workshop on Semantic Evaluation ({S}em{E}val-2016)}, pages 31--41, San Diego, California. Association for Computational Linguistics.

\bibitem[{Ni et~al.(2019)Ni, Li, and McAuley}]{ni-etal-2019-justifying}
Jianmo Ni, Jiacheng Li, and Julian McAuley. 2019.
\newblock \href {https://doi.org/10.18653/v1/D19-1018} {Justifying recommendations using distantly-labeled reviews and fine-grained aspects}.
\newblock In \emph{Proceedings of the 2019 Conference on Empirical Methods in Natural Language Processing and the 9th International Joint Conference on Natural Language Processing (EMNLP-IJCNLP)}, pages 188--197, Hong Kong, China. Association for Computational Linguistics.

\bibitem[{Olabisi et~al.(2022)Olabisi, Hudson, Jetter, and Agrawal}]{olabisi-etal-2022-analyzing}
Olubusayo Olabisi, Aaron Hudson, Antonie Jetter, and Ameeta Agrawal. 2022.
\newblock \href {https://aclanthology.org/2022.coling-1.542} {Analyzing the dialect diversity in multi-document summaries}.
\newblock In \emph{Proceedings of the 29th International Conference on Computational Linguistics}, pages 6208--6221, Gyeongju, Republic of Korea. International Committee on Computational Linguistics.

\bibitem[{Ouyang et~al.(2022)Ouyang, Wu, Jiang, Almeida, Wainwright, Mishkin, Zhang, Agarwal, Slama, Ray et~al.}]{ouyang2022training}
Long Ouyang, Jeffrey Wu, Xu~Jiang, Diogo Almeida, Carroll Wainwright, Pamela Mishkin, Chong Zhang, Sandhini Agarwal, Katarina Slama, Alex Ray, et~al. 2022.
\newblock Training language models to follow instructions with human feedback.
\newblock \emph{Advances in Neural Information Processing Systems}, 35:27730--27744.

\bibitem[{Rafailov et~al.(2024)Rafailov, Sharma, Mitchell, Manning, Ermon, and Finn}]{rafailov2024direct}
Rafael Rafailov, Archit Sharma, Eric Mitchell, Christopher~D Manning, Stefano Ermon, and Chelsea Finn. 2024.
\newblock Direct preference optimization: Your language model is secretly a reward model.
\newblock \emph{Advances in Neural Information Processing Systems}, 36.

\bibitem[{Raffel et~al.(2020)Raffel, Shazeer, Roberts, Lee, Narang, Matena, Zhou, Li, and Liu}]{raffel2020exploring}
Colin Raffel, Noam Shazeer, Adam Roberts, Katherine Lee, Sharan Narang, Michael Matena, Yanqi Zhou, Wei Li, and Peter~J Liu. 2020.
\newblock Exploring the limits of transfer learning with a unified text-to-text transformer.
\newblock \emph{Journal of machine learning research}, 21(140):1--67.

\bibitem[{Randolph(2005)}]{randolph2005free}
Justus~J Randolph. 2005.
\newblock Free-marginal multirater kappa (multirater k [free]): An alternative to fleiss' fixed-marginal multirater kappa.
\newblock \emph{Online submission}.

\bibitem[{Roit et~al.(2023)Roit, Ferret, Shani, Aharoni, Cideron, Dadashi, Geist, Girgin, Hussenot, Keller, Momchev, Ramos~Garea, Stanczyk, Vieillard, Bachem, Elidan, Hassidim, Pietquin, and Szpektor}]{roit-etal-2023-factually}
Paul Roit, Johan Ferret, Lior Shani, Roee Aharoni, Geoffrey Cideron, Robert Dadashi, Matthieu Geist, Sertan Girgin, Leonard Hussenot, Orgad Keller, Nikola Momchev, Sabela Ramos~Garea, Piotr Stanczyk, Nino Vieillard, Olivier Bachem, Gal Elidan, Avinatan Hassidim, Olivier Pietquin, and Idan Szpektor. 2023.
\newblock \href {https://doi.org/10.18653/v1/2023.acl-long.344} {Factually consistent summarization via reinforcement learning with textual entailment feedback}.
\newblock In \emph{Proceedings of the 61st Annual Meeting of the Association for Computational Linguistics (Volume 1: Long Papers)}, pages 6252--6272, Toronto, Canada. Association for Computational Linguistics.

\bibitem[{Shandilya et~al.(2018)Shandilya, Ghosh, and Ghosh}]{shandilya2018fairness}
Anurag Shandilya, Kripabandhu Ghosh, and Saptarshi Ghosh. 2018.
\newblock Fairness of extractive text summarization.
\newblock In \emph{Companion Proceedings of the The Web Conference 2018}, pages 97--98.

\bibitem[{Stiennon et~al.(2020)Stiennon, Ouyang, Wu, Ziegler, Lowe, Voss, Radford, Amodei, and Christiano}]{stiennon2020learning}
Nisan Stiennon, Long Ouyang, Jeffrey Wu, Daniel Ziegler, Ryan Lowe, Chelsea Voss, Alec Radford, Dario Amodei, and Paul~F Christiano. 2020.
\newblock Learning to summarize with human feedback.
\newblock \emph{Advances in Neural Information Processing Systems}, 33:3008--3021.

\bibitem[{Team et~al.(2024)Team, Mesnard, Hardin, Dadashi, Bhupatiraju, Pathak, Sifre, Rivi{\`e}re, Kale, Love et~al.}]{team2024gemma}
Gemma Team, Thomas Mesnard, Cassidy Hardin, Robert Dadashi, Surya Bhupatiraju, Shreya Pathak, Laurent Sifre, Morgane Rivi{\`e}re, Mihir~Sanjay Kale, Juliette Love, et~al. 2024.
\newblock Gemma: Open models based on gemini research and technology.
\newblock \emph{arXiv preprint arXiv:2403.08295}.

\bibitem[{Zhang et~al.(2023)Zhang, Zhang, Liu, Fabbri, Liu, Kamoi, Lu, Xiong, Zhao, Radev et~al.}]{zhang2023fair}
Yusen Zhang, Nan Zhang, Yixin Liu, Alexander Fabbri, Junru Liu, Ryo Kamoi, Xiaoxin Lu, Caiming Xiong, Jieyu Zhao, Dragomir Radev, et~al. 2023.
\newblock Fair abstractive summarization of diverse perspectives.
\newblock \emph{arXiv preprint arXiv:2311.07884}.

\bibitem[{Ziegler et~al.(2019)Ziegler, Stiennon, Wu, Brown, Radford, Amodei, Christiano, and Irving}]{ziegler2019fine}
Daniel~M Ziegler, Nisan Stiennon, Jeffrey Wu, Tom~B Brown, Alec Radford, Dario Amodei, Paul Christiano, and Geoffrey Irving. 2019.
\newblock Fine-tuning language models from human preferences.
\newblock \emph{arXiv preprint arXiv:1909.08593}.

\end{thebibliography}

\appendix

\section{Appendix}
\subsection{Datasets}
\label{sec:dataset}
In this section, we describe how we preprocess the datasets. 
\paragraph{Amazon} \cite{ni-etal-2019-justifying} consists of reviews with labels of their ratings of different products. We filter out reviews that are non-English or without ratings. We obtain the social attribute of each review based on its rating provided in the dataset. The social attribute of a review will be positive if its rating is $4$ or $5$, neutral if its rating is $3$, and negative if its rating is $1$ or $2$. To construct training, validation and testing sets, we perform stratified sampling based on the distribution of social attribute values among document sets for each set. Therefore, each set has equal proportions of document sets $D$ dominated by each social attribute values. We sample 1000 products and their corresponding reviews for training, 300 products for validation, and 300 products for testing. 

\paragraph{MITweet} \cite{liu2023ideology} consists of tweets with labels of political ideologies on different facets about different topics. The social attribute of a tweet will be left if it is left on most facets, right if it is right on most facets, otherwise neutral. First, we evenly divide all tweets of each topic into two parts so that the distribution of topics is the same between two parts. For each part, we cluster tweets about the same topic based on their TFIDF similarity into clusters. We then divide these clusters into input document sets of 20 tweets about the same topic. We generate 1000 input document sets for training from the first part of the tweets. Similarly, we generate 300 input document sets for validation and 300 input document sets for testing from the second part of the tweets. When generating input document sets of training, validation, and testing sets, we also perform stratified sampling based on the distribution of social attribute values so that each set has equal proportions of document sets $D$ dominated by each social attribute value.

\paragraph{Tweet Stance} \cite{mohammad-etal-2016-semeval} consists of tweets with labels of stance toward a target phrase such as Climate Change or Hillary Clinton. First, we evenly divide all tweets of each topic into two parts so that the distribution of target phrase is the same between two parts. We cluster tweets about the same target phrase based on their TFIDF similarity into clusters. We then divide these clusters into input document sets of 30 tweets about the same target phrase. We generate 1000 input document sets for training from the first part of the tweets. Similarly, we generate 300 input document sets for validation and 300 input document sets for testing from the second part of the tweets. When generating input document sets of training, validation, and testing sets, we also perform stratified sampling based on the distribution of social attribute values so that each set has equal proportions of document sets $D$ dominated by each social attribute value.

\subsection{Human Evaluation}
\label{app:human}
\begin{figure*}
\centering
\includegraphics[width=0.6\textwidth]{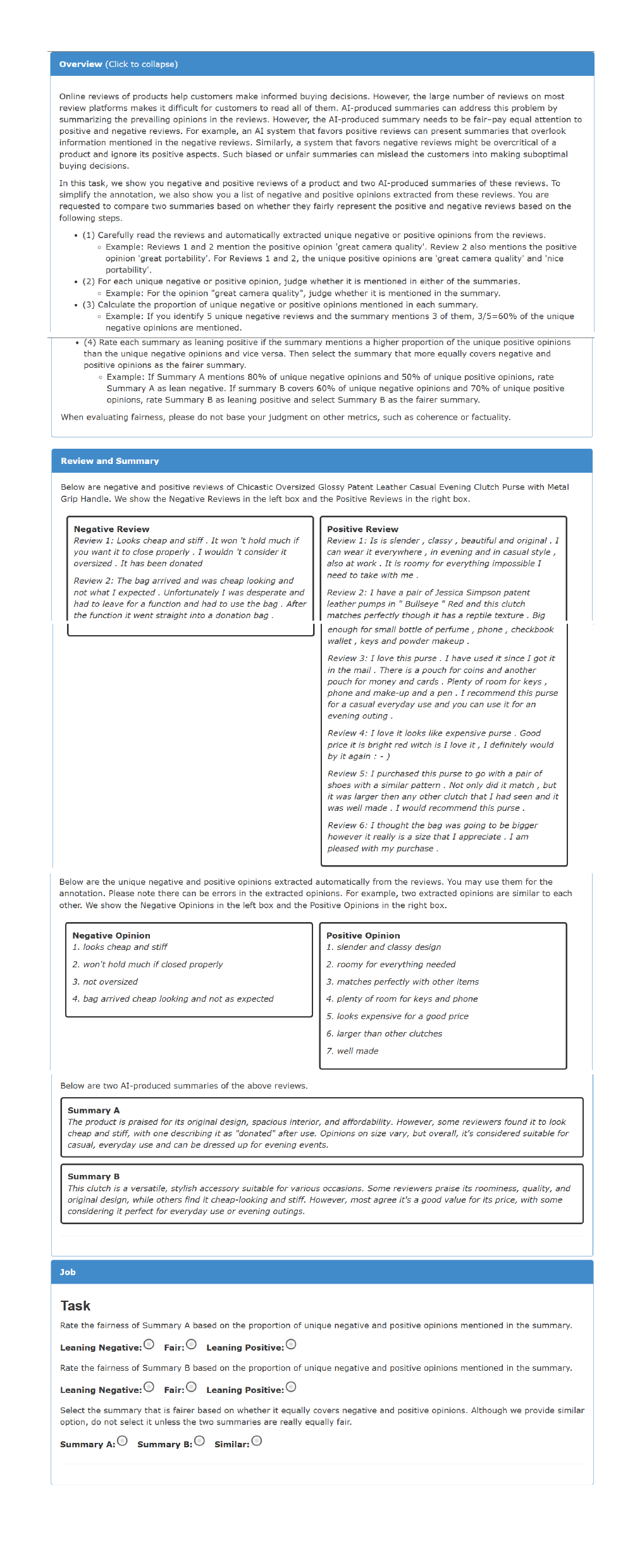}
\caption{Interface for Human Evaluation}
\label{fig:human}
\end{figure*}

We perform a human evaluation to compare the fairness of summaries generated by LLMs tuned with DPO and FairPO. For each LLM, we randomly select $10$ pairs of summaries generated by the LLM tuned with DPO or FairPO, yielding a total of $30$ pairs. To further simplify the evaluation, we consider document sets with only negative and positive reviews. Each pair is annotated by three annotators recruited from Amazon Mechanical Turk. The annotators should be from English-speaking countries and have HIT Approval Rates greater than $98\%$. For each pair, annotators are first asked to read corresponding reviews and unique opinions automatically extracted by GPT-4o-mini \cite{ouyang2022training}. They then evaluate whether each summary reflects these opinions and classify the summary as leaning negative, fair, or leaning positive. Eventually, they are asked to select the fairer summary in each pair. The interface of human evaluation is shown in Fig. \ref{fig:human}.  

\begin{figure*}
\centering
\includegraphics[width=0.95\textwidth]{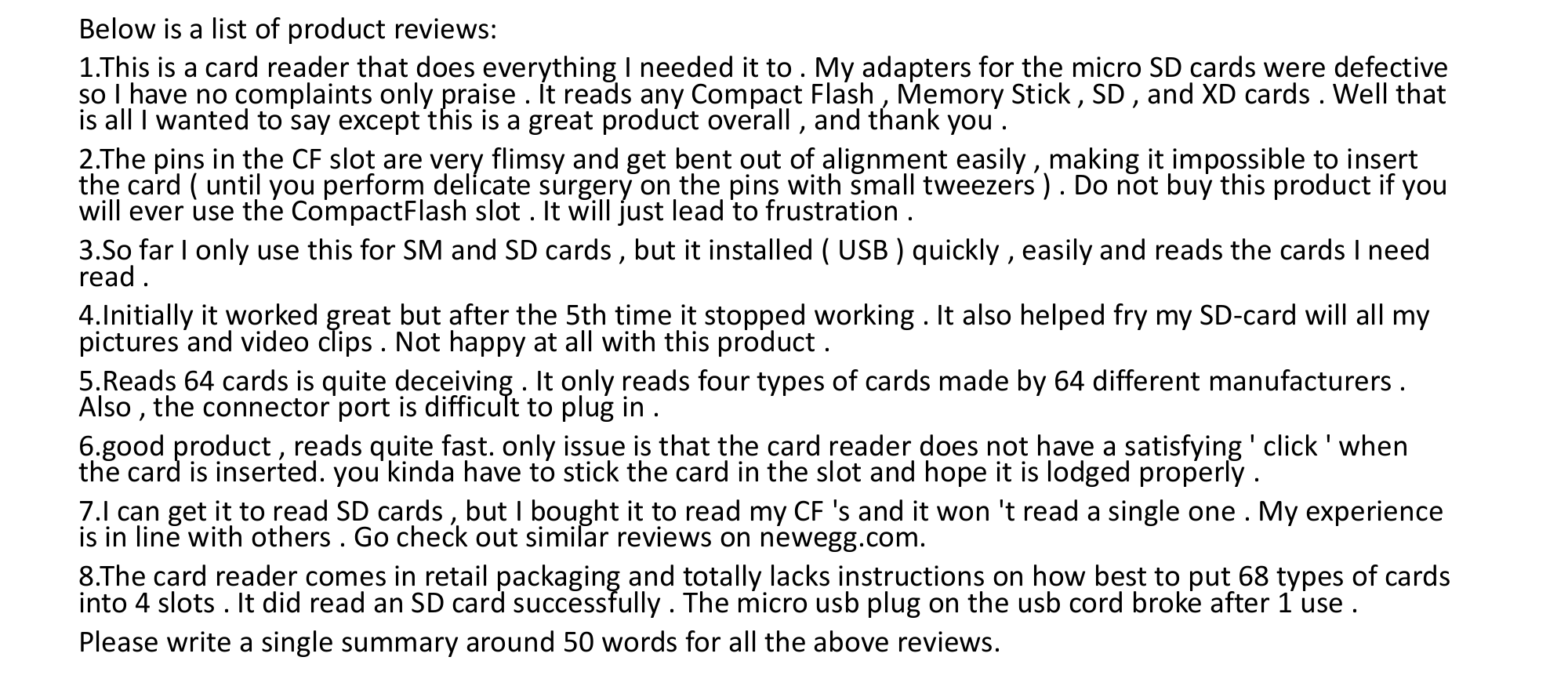}
\caption{Summarization prompt for the Amazon Dataset.}
\label{fig:sum_prompt}
\end{figure*}

\subsection{Relation between FairPO and DPO}
\label{sec:relation}
The FairPO objective (Eq. \ref{eq:objective}) is motivated by the derivate of the  DPO objective with respect to the model parameters $\theta$:
\begin{equation}
\begin{aligned}
\sigma(-m)\beta(\pi_\theta(S_r|D)^{-1}\frac{\partial\pi_{\theta}(S_r|D)}{\partial\theta}
\\-\pi_\theta(S_c|D)^{-1}\frac{\partial\pi_{\theta}(S_c|D)}{\partial\theta})
\end{aligned}
\label{eq:dpo}
\end{equation}
where $\sigma$ is the sigmoid function, $\pi_\theta$ is the policy model, $\pi_{ref}$ is the reference model, and $m$ is the reward margin in DPO:
\begin{equation}
\beta log\frac{\pi_\theta(S_c|D)}{\pi_{ref}(S_c|D)}-\beta log\frac{\pi_\theta(S_r|D)}{\pi_{ref}(S_r|D)}
\end{equation}
 The reward margin $m$ can be viewed as a measure of the model’s ability to distinguish between the chosen summary $S_c$ and the rejected summary $S_r$. A larger value of $m$ indicates that the model is already proficient at differentiating $S_c$ from $S_r$. Consequently, DPO assigns lower weights, $\sigma(-m)$, to chosen and rejected summaries where the model is confident in their differences and higher weights to chosen and rejected summaries where the differences is more challenging. The term $\sigma(-m)$ can help the model focuses more on difficult cases.

The objective of FairPO is designed so that chosen and rejected summaries have separate weight while preserving the effect of the term $\sigma(-m)$ in Eq.\ref{eq:dpo}. The derivative of FariPO objective with respect to the model parameters $\theta$ is as follows:
\begin{equation}
\begin{aligned}
\sigma(-m)\beta(w_r\pi_\theta(S_r|D)^{-1}\frac{\partial\pi_{\theta}(S_r|D)}{\partial\theta}
\\-w_c\pi_\theta(S_c|D)^{-1}\frac{\partial\pi_{\theta}(S_c|D)}{\partial\theta})
\end{aligned}
\end{equation}
Comparing with the derivative of DPO objective (Eq. \ref{eq:dpo}), the term $\sigma(-m)$ remains consistent in the derivative of FairPO objective. 

Suppose we directly add seperate weights $w_c$ and $w_r$ for chosen and rejected summaries to DPO objective. The corresponding objective is as follows:
\begin{equation}
\begin{aligned}
-log\sigma(\beta w_clog\frac{\pi_\theta(S_c|D)}{\pi_{ref}(S_c|D)}-\\
\beta w_r log\frac{\pi_\theta(S_r|D)}{\pi_{ref}(S_r|D)})
\end{aligned}
\label{eq:reward}
\end{equation}
The corresponding derivative is as follows:
\begin{equation}
\begin{aligned}
\sigma(-m')\beta(w_r\pi_\theta(S_r|D)^{-1}\frac{\partial\pi_{\theta}(S_r|D)}{\partial\theta}
\\-w_c\pi_\theta(S_c|D)^{-1}\frac{\partial\pi_{\theta}(S_c|D)}{\partial\theta})
\end{aligned}
\end{equation}
where $m'$ is a weighted reward margin:
\begin{equation}
\beta w_c log\frac{\pi_\theta(S_c|D)}{\pi_{ref}(S_c|D)}-\beta w_rlog\frac{\pi_\theta(S_r|D)}{\pi_{ref}(S_r|D)}
\end{equation}
Comparing with $m$, $m'$ is less effective as a measure of the model’s ability to distinguish between the chosen summary $S_c$ and the rejected summary $S_r$ since the term $log\frac{\pi_\theta(S_c|D)}{\pi_{ref}(S_c|D)}$ and $log\frac{\pi_\theta(S_r|D)}{\pi_{ref}(S_r|D)}$ have different weights. We additionally provide empirical evidences in App.\ref{sec:ablation}.

\subsection{Implementation Details}
\label{sec:implementation}
To reduce training cost, we perform LoRA \cite{hulora} tuning. Specifically, the rank for LoRA tuning is $16$ and the scaling factor is also $16$. All models are quantized in $8$-bit to additionally reducing training cost.

When performing perturbation on each document set to generate preference pairs, we observe that certain social attribute values are extremely rare in some document sets. If FairPO removes $\alpha$ percent of documents with these rare social attribute values, those social attribute values will disappear entirely from the document set. Therefore, when performing perturbation, we only consider social attribute values that appear in more than $\alpha$ percent of the documents. In the most extreme case, if only one social attribute value meets this requirement, FairPO will sample different subsets of $\alpha$ percent of documents with that social attribute value. By doing this, we assure the completeness of social attribute values after perturbation.

We prompt these LLMs to generate summaries for the input document sets of different datasets. The prompt are tuned so that the average length of generated summaries are $50$ words. We show the summarization prompts for the Amazon dataset in Fig. \ref{fig:sum_prompt}. The temperature for generation is $0.6$ for all LLMs.

The set $T_k^+$ in Eq.\ref{eq:over} is updated so that recent training steps have higher impacts. Specifically, at the end of each training step, the impacts of all the samples already in the set $T_k^+$ are reduced with a discount factor $\gamma$. Then, all the samples that overrepresents social attribute value $k$ ($C_k(D,S_*)$>0) in current training steps are added to the set $T_k^+$. The discount factor $\gamma$ is $0.75$ for Llama3.1 and $0.5$ for other LLMs.

The goal of the exponent, $C_k(D,S_*)$, of $O(k)/(U(k)$ or $U(k)/(O(k)$ in Eq. \ref{eq:cond1} and Eq. \ref{eq:cond2} is to adjust the weights $w_c$ and $w_r$ such that it more deviates from $1$ as $C_k(D,S_*)$ more deviates from $0$. Therefore, FairPO does not directly use the raw value of the sum of coverage probability differences $C_k(D,S_*)$ as the exponent. Instead, FairPO separately normalizes $C_k(D,S_*)$ among all training samples where $C_k(D,S_*)$ is greater than zero or less than zero.

\subsection{Implementation of Baseline}
\label{sec:baseline}
We implement the policy gradient method proposed by \citet{lei2024polarity} as a baseline. In the original implementation, there is a loss that maximize the probability for reference summary in addition to the policy gradients. Since datasets used in this paper do not contain reference summary, we only consider the policy gradients. Besides, for a fair comparison with other methods, we implement the policy gradient method in an offline setting. The learning rate for the policy gradeint is $1e-6$ following the original paper. We only implement the policy gradient method for Llama3.1 since the training is very unstable even if we lower the learning rate to $1e-9$ for Mistral and Gemma2. For OPTune and DPO, they use the same hyperparameters as FairPO.

\subsection{Results using Different Dataset Splitting}
\label{sec:split}
\begin{table}[]
\setlength{\tabcolsep}{1mm}
\resizebox{0.48\textwidth}{!}{
\begin{tabular}{lcccccccc}
\hline

                 & \multicolumn{2}{c}{Amazon}                & \multicolumn{2}{c}{MITweet}         & \multicolumn{2}{c}{SemEval}         &  \multicolumn{2}{c}{Overall}               \\
                 & $EC\downarrow$                  & $CP\downarrow$                  & $EC$                  & $CP\downarrow$            & $EC\downarrow$                  & $CP\downarrow$            & $\overline{EC}\downarrow$           & $\overline{CP}\downarrow$            \\ \hline
Llama3.1   & 7.90          & 1.92          & 4.43          & 0.26          & 2.94          & 1.33          & 5.09          & 1.17          \\
~+DPO       & 6.87          & 1.04          & \textbf{4.03} & 0.31          & 2.55          & 0.91          & 4.49          & 0.75          \\
~+OPTune    & \underline{6.58} & \underline{0.75} & 4.22          & \textbf{0.23} & \underline{2.50}          & 0.81          & \underline{4.43} & \underline{0.60} \\
~+Prompt    & 7.71          & 1.84          & 4.33          & 0.38          & 2.53          & \textbf{0.26} & 4.86          & 0.83          \\
~+Policy G. & 7.71          & 2.10          & 4.46          & 0.31          & 2.95          & 1.32          & 5.04          & 1.24          \\
~+FairPO    & \textbf{6.57} & \textbf{0.37} & \underline{4.20} & \underline{0.26} & \textbf{2.39} & \underline{0.56} & \textbf{4.39} & \textbf{0.39} \\ \hdashline
Mistral    & 8.18          & 2.98          & 3.98          & 0.42          & 2.67          & 1.07          & 4.94          & 1.49          \\
~+DPO       & \underline{7.17} & \underline{1.55} & \underline{3.60} & 0.28          & 2.21          & 0.64          & \underline{4.33} & \underline{0.82} \\
~+OPTune    & 7.48          & 1.56          & 3.60          & 0.25          & \underline{2.00} & 0.67          & 4.36          & 0.83          \\
~+Prompt    & 7.67          & 1.93          & 4.02          & \underline{0.23} & 2.38          & \underline{0.38} & 4.69          & 0.85          \\
~+FairPO    & \textbf{6.98} & \textbf{0.89} & \textbf{3.56} & \textbf{0.21} & \textbf{1.97} & \textbf{0.36} & \textbf{4.17} & \textbf{0.49} \\ \hdashline
Gemma2     & 8.44          & 2.75          & 4.17          & 0.34          & 2.74          & 0.91          & 5.12          & 1.33          \\
~+DPO       & \underline{6.87} & \underline{1.04} & 4.04          & \underline{0.29} & \underline{2.42} & 0.70          & 4.44          & 0.68          \\
~+OPTune    & 6.90          & 1.15          & \underline{3.86}          & 0.45          & \textbf{2.40} & 0.65          & \underline{4.39} & 0.75          \\
~+Prompt    & 7.21          & 1.13          & 4.28          & \textbf{0.24} & 2.62          & \textbf{0.30} & 4.70          & \underline{0.56} \\
~+FairPO    & \textbf{6.09} & \textbf{0.33} & \textbf{3.84} & 0.47          & 2.53          & \underline{0.59} & \textbf{4.15} & \textbf{0.46} \\
 \hline
\end{tabular}}
\caption{Summary-level fairness ($EC$) and corpus-level fairness ($CP$) of summaries generated by different methods on the first splitting of training, validation and testing.. The best performing method is in \textbf{bold}. The second-best performing method is \underline{underlined}. FairPO has the best overall performance on the first splitting.}
\label{tab:split1}
\end{table}

\begin{table}[t!]
\setlength{\tabcolsep}{1mm}
\resizebox{0.48\textwidth}{!}{
\begin{tabular}{lcccccccc}
\hline

                 & \multicolumn{2}{c}{Amazon}                & \multicolumn{2}{c}{MITweet}         & \multicolumn{2}{c}{SemEval}         &  \multicolumn{2}{c}{Overall}               \\
                 & $EC\downarrow$                  & $CP\downarrow$                  & $EC$                  & $CP\downarrow$            & $EC\downarrow$                  & $CP\downarrow$            & $\overline{EC}\downarrow$           & $\overline{CP}\downarrow$            \\ \hline
Llama3.1 & 7.90          & 2.05          & 4.50          & 0.63          & 2.90          & 1.41          & 5.10          & 1.36          \\
~+DPO     & 7.27          & 1.37          & {\ul 4.30}    & {\ul 0.37}    & 2.70          & 1.12          & 4.76          & 0.95          \\
~+OPTune  & \textbf{6.92} & \textbf{0.40} & 4.30          & 0.52          & 2.82          & 1.00          & {\ul 4.68}    & {\ul 0.64}    \\
~+Prompt  & 7.28          & 1.67          & 4.41          & 0.44          & {\ul 2.74}    & \textbf{0.51} & 4.81          & 0.87          \\
~+Policy G. & 7.75 & 1.85 & 4.47 & 0.48 & 2.80 & 1.30 & 5.02 & 1.21 \\
~+FairPO  & {\ul 6.96}    & {\ul 0.44}    & \textbf{4.26} & \textbf{0.29} & \textbf{2.69} & {\ul 0.59}    & \textbf{4.64} & \textbf{0.44} \\ \hdashline
Mistral  & 8.60          & 2.74          & 4.18          & 0.73          & 2.91          & 1.28          & 5.23          & 1.58          \\
~+DPO     & 7.24          & 1.79          & \textbf{3.39} & \textbf{0.26} & 2.70          & 1.15          & 4.44          & 1.07          \\
~+OPTune  & {\ul 6.59}    & {\ul 0.52}    & \textbf{3.57} & 0.53          & \textbf{2.04} & 0.58          & {\ul 4.07}    & {\ul 0.54}    \\
~+Prompt  & 7.90          & 1.76          & 3.74          & 0.51          & 2.43          & 0.52          & 4.69          & 0.93          \\
~+FairPO  & \textbf{6.06} & \textbf{0.11} & 3.83          & {\ul 0.39}    & {\ul 2.13}    & \textbf{0.33} & \textbf{4.01} & \textbf{0.28} \\ \hdashline
Gemma2   & 8.31          & 2.33          & 4.30          & 0.80          & 2.97          & 1.03          & 5.19          & 1.38          \\
~+DPO     & 7.04          & 0.98          & 4.07          & {\ul 0.44}    & {\ul 2.43}    & {\ul 0.48}    & 4.51          & {\ul 0.63}    \\
~+OPTune  & {\ul 6.91}    & {\ul 0.56}    & 3.94          & 0.86          & \textbf{2.35} & 0.56          & {\ul 4.40}    & 0.66          \\
~+Prompt  & 7.33          & 1.26          & 4.49          & \textbf{0.44} & 2.91          & 0.85          & 4.91          & 0.85          \\
~+FairPO  & \textbf{6.09} & \textbf{0.44} & \textbf{3.82} & 0.65          & 2.70          & \textbf{0.32} & \textbf{4.20} & \textbf{0.47} \\
 \hline
\end{tabular}}
\caption{Summary-level fairness ($EC$) and corpus-level fairness ($CP$) of summaries generated by different methods on the second splitting of training, validation and testing. The best performing method is in \textbf{bold}. The second-best performing method is \underline{underlined}. FairPO has the best overall performance on the second splitting.}
\label{tab:split2}
\end{table}

\begin{table}[t!]
\setlength{\tabcolsep}{1mm}
\resizebox{0.48\textwidth}{!}{
\begin{tabular}{lcccccccc}
\hline

                 & \multicolumn{2}{c}{Amazon}                & \multicolumn{2}{c}{MITweet}         & \multicolumn{2}{c}{SemEval}         &  \multicolumn{2}{c}{Overall}               \\
                 & $EC\downarrow$                  & $CP\downarrow$                  & $EC$                  & $CP\downarrow$            & $EC\downarrow$                  & $CP\downarrow$            & $\overline{EC}\downarrow$           & $\overline{CP}\downarrow$            \\ \hline
Llama3.1   & 8.06          & 1.70          & 4.57          & 0.87          & 3.11          & 1.51          & 5.25          & 1.36          \\
+DPO       & 7.55          & 1.39          & 4.43          & 0.74          & 2.73          & 1.24          & 4.90          & 1.12          \\
+OPTune    & \textbf{6.61} & {\ul 0.72}    & 4.47          & 0.79          & {\ul 2.48}    & 1.03          & \textbf{4.52} & 0.85          \\
+Prompt    & 7.26          & 1.42          & {\ul 4.35}    & \textbf{0.53} & 2.59          & \textbf{0.11} & 4.74          & {\ul 0.69}    \\
+Policy G. & 7.72          & 1.69          & 4.60          & 0.85          & 3.16          & 1.53          & 5.16          & 1.36          \\
+FairPO    & {\ul 7.07}    & \textbf{0.44} & \textbf{4.25} & {\ul 0.71}    & \textbf{2.38} & {\ul 0.82}    & {\ul 4.57}    & \textbf{0.66} \\ \hdashline
Mistral    & 8.29          & 2.76          & 4.32          & 0.67          & 2.90          & 1.46          & 5.17          & 1.63          \\
+DPO       & 7.20          & 2.11          & {\ul 3.65}    & {\ul 0.48}    & 2.32          & 0.99          & 4.39          & 1.19          \\
+OPTune    & {\ul 6.47}    & {\ul 0.55}    & \textbf{3.58} & 0.76          & \textbf{2.18} & {\ul 0.46}    & {\ul 4.08}    & {\ul 0.59}    \\
+Prompt    & 7.66          & 2.07          & 4.14          & \textbf{0.36} & 2.23          & \textbf{0.19} & 4.68          & 0.87          \\
+FairPO    & \textbf{5.92} & \textbf{0.38} & 3.71          & 0.61          & {\ul 2.21}    & 0.59          & \textbf{3.95} & \textbf{0.53} \\ \hdashline
Gemma2     & 8.21          & 2.36          & 4.14          & 0.67          & 2.72          & 0.96          & 5.02          & 1.33          \\
+DPO       & 6.80          & {\ul 0.70}    & 4.01          & 0.49          & 2.46          & 0.49          & 4.42          & 0.56          \\
+OPTune    & {\ul 6.72}    & 0.94          & {\ul 3.86}    & 0.41          & \textbf{2.21} & \textbf{0.25} & {\ul 4.27}    & {\ul 0.53}    \\
+Prompt    & 7.29          & 1.09          & 4.21          & \textbf{0.28} & 2.66          & {\ul 0.28}    & 4.72          & 0.55          \\
+FairPO    & \textbf{6.35} & \textbf{0.56} & \textbf{3.64} & {\ul 0.32}    & {\ul 2.29}    & 0.43          & \textbf{4.09} & \textbf{0.44} \\
 \hline

\end{tabular}}
\caption{Summary-level fairness ($EC$) and corpus-level fairness ($CP$) of summaries generated by different methods on the third splitting of training, validation and testing. The best performing method is in \textbf{bold}. The second-best performing method is \underline{underlined}. FairPO has the best overall performance on the third splitting.}
\label{tab:split3}
\end{table}

To validate the stability of FairPO on three different splittings of datasets, we generate the training, validation and testing set using different random seeds and run the automatic evaluation on these splittings. The results for each splitting are shown in Tab. \ref{tab:split1}, \ref{tab:split2}, \ref{tab:split3} respectively. From the table, we observe that FairPO still shows the best overall performance, which shows the stability of  FairPO on different splittings of datasets.

\subsection{Qualiative Example}
\label{sec:qualitative}
\begin{figure*}
\centering
\includegraphics[width=0.95\textwidth]{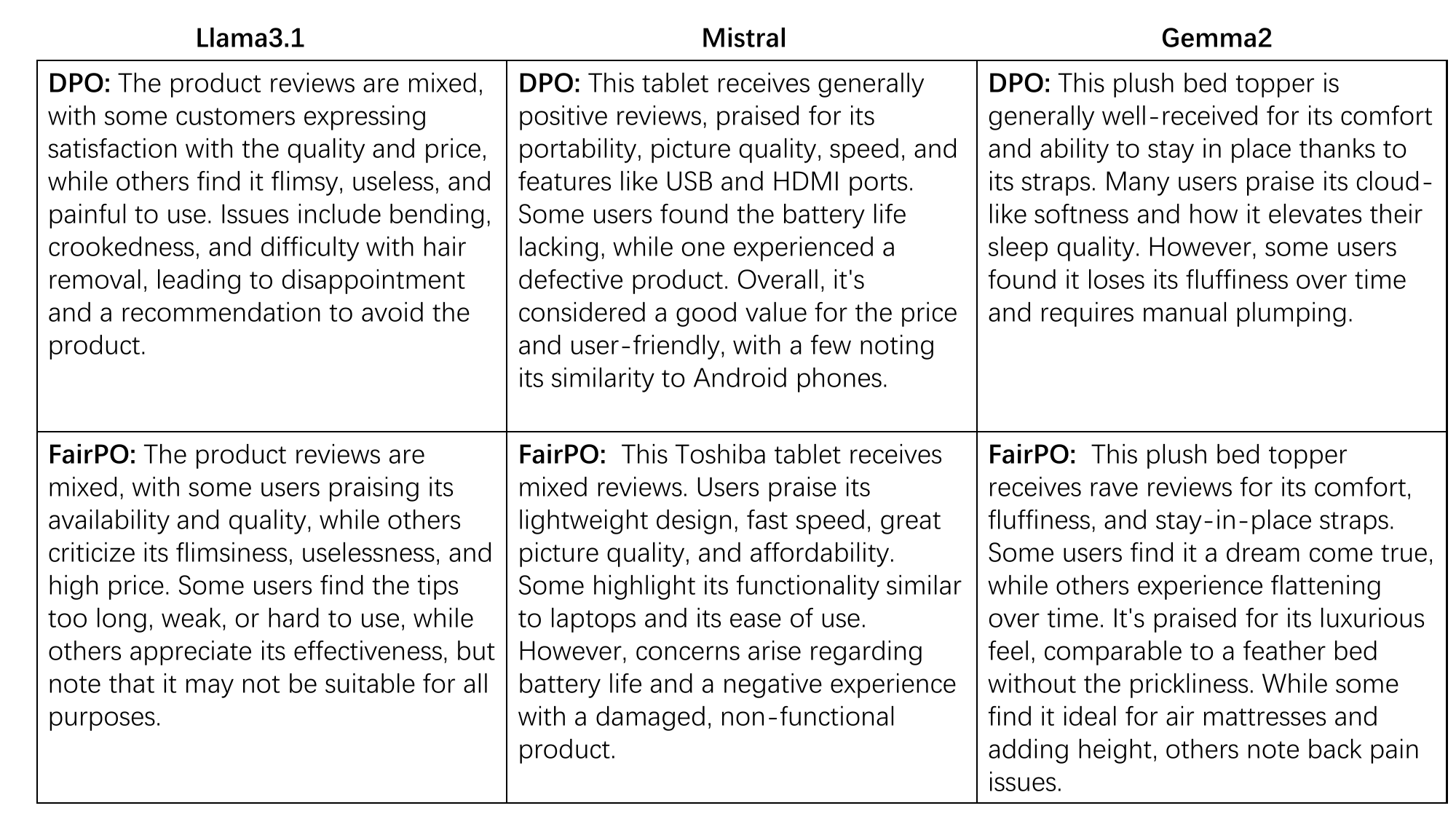}
\caption{Sample summaries generated by DPO and FairPO.}
\label{fig:sample}
\end{figure*} 
We show sample summaries generated by LLMs tuned with DPO and FairPO on the Amazon dataset in Fig. \ref{fig:sample}. From the figure, we observe that summaries generated by LLMs tuned FairPO tend to more balancely present negative and positive information.

\end{document}